\documentclass[final,5p,twocolumn]{elsarticle}
\usepackage{newtxtext}

\usepackage{amssymb}
\usepackage{amsmath}

\usepackage[caption=false,font=normalsize,labelfont=sf,textfont=sf]{subfig}
\usepackage{xcolor}
\usepackage{amsfonts}
\usepackage{amsthm}
\usepackage{algorithm2e} 
\SetAlCapNameFnt{\footnotesize}
\SetAlCapFnt{\footnotesize}
\SetAlgoCaptionLayout{raggedright}
\SetAlFnt{\footnotesize}

\usepackage{booktabs}
\usepackage{mathrsfs}
\usepackage{hyperref} 
\usepackage{multirow}

\definecolor{color_congruent}{rgb}{0., 0., 1.}
\definecolor{color_concentric}{rgb}{0. , 0.50196078, 0.74901961}
\definecolor{color_screwed}{rgb}{0. , 1. , 0.5}

\newcommand\frameworkacr{GLARE}
\newcommand\frameworkname{\textbf{G}eometric-symbolic \textbf{L}earning-enabled \textbf{A}ssembly \textbf{R}easoning \textbf{E}ngine}
\newcommand{\sqboxs}{1.2ex}
\newcommand{\sqbox}[1]{\textcolor{#1}{\rule{\sqboxs}{\sqboxs}}}
\newcommand{\sqboxblack}[1]{\setlength{\fboxsep}{0pt}\fbox{\sqbox{#1}}}

\journal{Robotics and Computer-Integrated Manufacturing}

\begin{document}

\begin{frontmatter}



\title{CAD-Based Relation Learning and Geometric-Symbolic Planning for Robotic Assembly \newline}


\cortext[cor]{Corresponding author, \textit{Email adress}: christian.friedrich@hka.de}

\author[HKA]{Fabian Harlacher} 
\author[HKA]{Christian Friedrich\corref{cor}} 

\affiliation[HKA]{organization={Institute for Robotics and Intelligent Production Systems, Karlsruhe University of Applied Sciences (HKA)},
            addressline={Moltkestraße 30},
            city={Karlsruhe},
            postcode={76133}, 
            state={Baden-Württemberg},
            country={Germany}}

\begin{abstract}
Assembly Sequence Planning (ASP) remains a challenging problem due to its combinatorial nature, making exhaustive planning approaches impractical for complex industrial assemblies. Furthermore, many CAD models lack reliable semantic contact information or require extensive manual preprocessing, limiting the applicability of existing methods. This paper presents a hybrid ASP framework combining learning-based relation extraction with geometric-symbolic reasoning to generate feasible robotic disassembly sequences from imperfect CAD data. A neural network predicts semantic geometric relations from point clouds, while human-in-the-loop verification enables correction of uncertain predictions and planning failures. Extracted relations are transformed into a symbolic assembly graph, enabling a geometric-symbolic planner to efficiently compute locally valid sets of robotic manipulation primitives. A visibility-based ray-casting strategy guides the search for feasible disassembly directions without requiring an exhaustive combinatorial search, while the local solution space enables efficient sequence optimization. The proposed framework is evaluated on an introduced assembly dataset and on the ASAP test dataset. On the ASAP test dataset, the proposed planner achieves an 85.83\% planning success rate while reducing the median planning time by more than one order of magnitude across all assembly sizes and by more than a factor of 50 for assemblies with more than 30 components compared to the baseline. The results demonstrate that the proposed hybrid framework enables efficient robotic assembly sequence planning from imperfect CAD data while substantially reducing planning time. By combining learning-based feature segmentation, human-in-the-loop verification, and geometric-symbolic reasoning, the framework provides a practical foundation for scalable and adaptable robotic assembly and disassembly planning.
\end{abstract}




\end{frontmatter}



\section{Introduction}
Assembly planning is a fundamental task in robotics, as it enables the efficient automation of manufacturing and recycling tasks, even for products with low batch sizes and high variant diversity. For a robot system to perform this task, a suitable sequence of assembly operations must be determined through the Assembly Sequence Planning (ASP) process. The planner operates on geometric and/or semantic information describing the assembly, obtained for example from CAD models or camera data, and transforms this information into corresponding robotic manipulation primitives. Due to the NP-completeness of the ASP problem \cite{Kavraki1995}, it is necessary to design suitable heuristics to determine a semi-optimal assembly sequence in a timely manner. Usually, most methods choose an assembly-by-disassembly strategy, as this reduces the combinatorial complexity. It was shown that the combinatorial complexity of the disassembly space $ Z $ ranges between $ n(n^{2} - 1)/6 \leq \ Z \leq \ (3^{n} + 1)/2 - 2^{n} $, depending on weak connected groups (every component $ n $ has a maximum of two connections) or strong connected groups (every component $ n $ is connected to all others) \cite{Thomas2008}. Today, geometric and semantic information in the form of CAD data is mostly available for all products to be manufactured. However, this data is not usually utilized comprehensively for the assembly process, as it is often incomplete or the information from the CAD is difficult to transfer and interpret for the robot system. There exist interesting learning-based approaches, like \cite{Kienle2025} which is a LLM-based question-answer system to extract precise information which can be further used for automatic robot program parameterization. Such learning-based strategies are especially suited to overcome the limitations of different design tolerances and not-fully specified symbolic spatial relations. On the other hand, there exist computational efficient planners like \cite{Friedrich2018}, based on symbolic-geometric methods, which, however, require a fully-defined geometric and semantic description (relational assembly model) of the assembly group. To overcome these limitations, we propose \frameworkacr{} (\frameworkname), a novel planning framework that integrates hybrid reasoning with a combined learning-based and symbolic-geometric approach to achieve time-efficient and semi-optimal assembly planning. Based on our Semantic Feature Learning module, we can semi-automatically predict a fully defined relational assembly model, which is checked for completeness during the planning process and can be manually annotated by the user if necessary. This is the basis for our Geometric Disassembly and Symbolic Sequence Planner, which enables us to generate a series of symbolic instructions, like manipulation primitives, in a semi-optimal manner to perform the assembly task.

Our work focuses on the gap between data-driven (learning-based) and geometric-symbolic (feature-based) approaches to expand planning capabilities for complex real-world problems. This hybrid approach is robust against uncertainties in CAD data, like missing semantic contact information or geometric tolerances, while the symbolic-geometric planning core enables transparent and algorithmically efficient sequences to be determined. The main s of our work are:
\begin{itemize}
    \item To the best of our knowledge, we are introducing the first hybrid assembly planning approach that combines learning and symbolic-geometric planning. This enables transparency in the planning process, which is essential for many real world manufacturing applications.
    \item A time efficient approach that runs in linear-logarithmic time and ensures a semi-optimal solution with regard to the robot execution time of the planned assembly sequence, considering motion and tool changing costs.
    \item Automatic learning from relational assembly models based on purely geometric CAD data also under varying geometric tolerance limits.
    \item Automatic fault detection in the Geometric Disassembly Planner and the possibility of correcting them easily using human-in-the-loop annotation.
    \item A new dataset containing academic and real-world industrial assembly groups, which can easily be extended using our human-in-the-loop annotation tool to foster new ASP algorithms for geometric-symbolic or learning-based methods.
\end{itemize}
Finally, we perform a comprehensive evaluation of our overall approach, comparing to an existing state of the art method. We hope that this work will serve to further advance the research of hybrid approaches combining data-driven and symbolic-geometric methods in the field of robotic assembly planning.

\section{Related Work}

\noindent Assembly planning is a well-researched area within robotics. Early research based on geometric-symbolic methods has made great advances
in the planning of assembly sequences \cite{Thomas2010}, also for real-world problems. One of the main problems with this so-called feature-based approach is that suitable features are required that are defined according to the rules in the underlying planning data. Due to advances in deep learning, these methods have also been applied to assembly planning in order to overcome the challenge of underlying data specification, based on a more generally applicable approach \cite{Tian2024}. However, these data-driven methods provide no transparency regarding the individual planning steps and require a lot of data for generalization. In addition, the annotation of training data is very time-consuming. To date, most of these approaches have only been validated for simple, block-based or academic assembly groups, as a proof of principle. In addition, we will discuss the most relevant research for this work.

\subsection{Feature-based assembly planning}

These methods usually have in common that they determine collision-free component sequences with associated assembly directions based on the relational assembly model that can be derived from CAD data with predefined design rules. To calculate the collision-free disassembly direction, symbolic spatial relations such as congruent, concentric, or screwed are taken into account, which define the connection between geometric features (e.g., plane, line, etc.) of the assembled components. In addition to such relational representations, CAD-based approaches can directly exploit geometric information to determine whether components are blocked in a particular direction. For example, interference matrices can be used to represent the geometric feasibility of component movements, while precedence matrices encode the resulting disassembly relationships and sequences \cite{Prioli2022}. In \cite{Thomas2003}, the configuration space obstacles of the individual components are determined and geometrically possible joint directions are projected onto a circumscribing circle using stereographic projection. The work \cite{Friedrich2016} developed a rule-based approach that makes it possible to calculate the various translational disassembly directions. In \cite{Friedrich2018}, a sampling-based method is proposed that makes it possible to determine the different directions and the corresponding sequence based on an optimization approach. The work was enhanced in \cite{Friedrich2022}, where a method is proposed to decompose the symbolic plan into executable reactive robot manipulations. The relational assembly model proposed in \cite{Mosemann2000} serves as the basis. However, they consider a mixed environment model that contains CAD and camera data, as the algorithms cannot generate valid sequences from geometric data alone, since the prerequisite is a complete relational assembly model. Further, the system cannot deal with geometric tolerances which are usual in real-world assemblies. In addition, path planners are also applied, compare \cite{Cortes2008}. In \cite{Tian2022}, a physics-based approach is proposed. The idea is that each component receives a signed distance field, which is used to calculate collision information and the corresponding contact forces, which, in turn, are used to construct a reduced search space. However, the computational complexity is between $ \mathcal{O} (n^2) $ for sequential disassembly and $ \mathcal{O}(n!) $ for subassemblies or interlocking assemblies for $ n $-components.

\begin{figure*}[!ht]
    \centering
    \includegraphics[width=\textwidth]{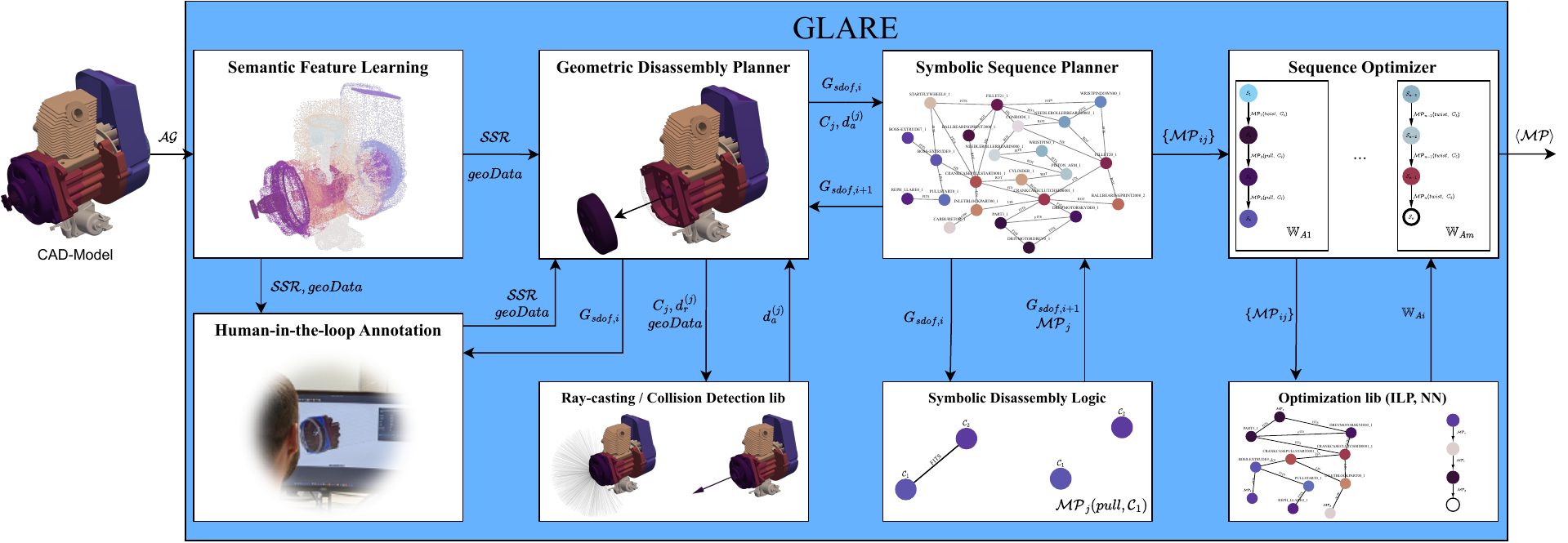}
    \caption{Framework architecture of \frameworkacr. The \textit{Semantic Feature Learning} module predicts the semantic spatial relations $\mathcal{SSR}$, which can be verified and corrected by the \textit{Human-in-the-Loop Annotation} module. The resulting symbolic relation graph $G_{ssr}$ is processed by the \textit{Geometric Disassembly Planner}, which combines ray-casting and collision checking to determine visibility spaces and valid absolute disassembly directions $d_a^{(j)}$. Based on these results, the \textit{Symbolic Sequence Planner} applies symbolic disassembly logic to generate sets of valid manipulation primitives $\{\mathcal{MP}_{ij}\}$. Finally, the \textit{Sequence Optimizer} computes the optimized manipulation sequence $\langle\mathcal{MP}\rangle$. In case of planning failure, the predicted relations can be re-evaluated and corrected through the human-in-the-loop annotation process.
    }
    \label{fig:figure_1_framework_architecture}
\end{figure*}

\subsection{Learning-based assembly planning}

Learning-based approaches also commonly build on the idea of the assembly-by-disassembly strategy. Starting from this, the next component to be disassembled is predicted. Deep reinforcement learning (DRL) is very common for this kind of application. In \cite{Neves2022} state-of-the-art DRL is applied to the ASP problem. Based on the stereographical projections from \cite{Thomas2003} an artificial neural network is trained to approximate the cost function to guide the search algorithm \cite{Kitz2021}. An interesting solution is provided in \cite{Liu2025} where a learning-based approach is combined with a physics-aware action mask to guide the RL agent. In \cite{Funk2022} they propose a GNN to represent the learned policy. The use of GNNs seems very suitable, as it allows for a straightforward mapping from the topological structure of the assemblies, as pursued in \cite{Cebulla2023}, \cite{Atad2023}. Both approaches are evaluated on aluminum profiles and do not consider complex assembly groups. In contrast, \cite{Tian2024} uses data from the Fusion Gallery Assembly Dataset \cite{Willis2022} and achieves a maximum success rate of up to 82\%. In \cite{Tie2025}, the use of assembly manuals is suggested, which, together with real camera images, are the input of a vision language model, which generates a hierarchical assembly graph. Other approaches develop a graph-transformer architecture which is evaluated by planning assembly sequences for building bricks \cite{Ma2022}. The work \cite{Xu2025} introduce the SPAformer model which is able to reduce the solution space based on assembly sequences. For evaluation, they build on an extended version of the PartNet benchmark \cite{Mo2019}. Beyond learning individual sequence decisions, recent approaches combine learned policies with explicit task-level planning. In \cite{Qi2026}, visual-language models extract subgoals from human demonstrations, which are converted into symbolic states by a large language model. A PDDL-based planner then generates the high-level action sequence, which is mapped to executable manipulation trajectories using a learned 3D diffusion policy. A similar hierarchical approach is presented in \cite{Wang2026}, where an LLM-based planning layer selects skills from a learned assembly skill library. The individual skills are learned from demonstrations using a Diffusion Policy and incorporate visual and force/torque information for closed-loop execution.

Consistent benchmarks and data are crucial for learning and evaluating the various approaches. Here, the use of simulation is essential for data generation. In \cite{Narang2022}, this is achieved by integrating a range of physical methods into a simulation and learning environment for contact-rich applications. For real-world transfer, it is essential to incorporate quality metrics that consider contact force energy and smoothness \cite{Schempp2025}.

\section{A Hybrid Approach For Assembly Planning}

\noindent The following section introduces the design of our \frameworkacr{} assembly planner. Due to the computational advantage, the assembly-by-disassembly strategy is employed. First, we introduce our framework architecture and semi-automatic learning process to receive a complete relational assembly model. In this preprocessing step, we also rely on supervised human-in-the-loop annotation in case the planner detects failures during the ASP process based on collision analysis. Second, the core algorithms for the geometric-symbolic sequence planning are introduced, which are composed into the \textit{Geometric Disassembly} and the \textit{Symbolic Sequence Planner}. Finally, we analyze the time complexity and optimality of the provided solution.

\begin{figure*}[!ht]
    \centering
    \subfloat[]{\includegraphics[width=\textwidth]{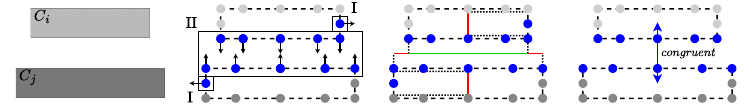}}%
    \label{fig:figure_2a_congruent_filtering}
    \subfloat[]{\includegraphics[width=\textwidth]{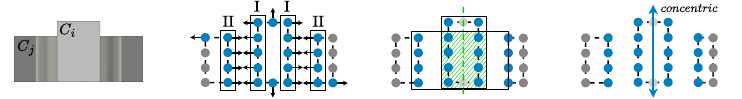}}%
    \label{fig:figure_2b_concentric_filtering}
    \caption{Determination of valid relations using two sliced parts and their semantic annotation from PointNet++ per approach as an example. (a) shows the procedure for $congruent$. Groups of labeled points are gathered by mapping to planar faces. Pairs of these groups with opposing normals ($\text{I}\text{ and }\text{II}$) are tested by overlap of the projection (green) of their respective faces to validate a relation. (b) shows the procedure for $concentric$ and $screwed$. Labeled points are also clustered with normals, but with an angle deviation, and further filtered by mapping back to cylindrical faces ($\text{I}\text{ and }\text{II}$). Further evaluation by axis alignment and overlapping of oriented bounding boxes (shaded green) validates a relation.}
    \label{fig:figure_2_filtering}
\end{figure*}

\subsection{Framework architecture and implementation}

\frameworkacr{} is modularized into five main cores as shown in \autoref{fig:figure_1_framework_architecture}. First, relations between the components are automatically extracted using \textit{Semantic Feature Learning}. These can be verified and corrected during the \textit{Human-in-the-loop Annotation}. The \textit{Geometric Disassembly Planner} provides a collision-free path for each component in the current assembly state. With this information, the \textit{Symbolic Sequence Planner} generates a sequence of valid actions to disassemble the corresponding components. Then, it selects a component to disassemble by removing it from the graph. The updated graph is then handed back to the \textit{Geometric Disassembly Planner}. This process continues until the assembly is fully disassembled. If no collision-free path is present, the current assembly state can be manually re-annotated. Finally, the \textit{Sequence Optimizer} generates an optimized sequence based on the sets of valid actions considering robot motion and tool changing costs. The procedure is described in detail in the following sections.

We implement \frameworkacr{} with Fusion360 \cite{Fusion360} for the \textit{Human-in-the-loop Annotation} and Python for the \textit{Semantic Feature Learning}, \textit{Geometric-Disassembly Planner}, \textit{Symbolic Sequence Planner} and \textit{Sequence Optimizer}. For creating and manipulating graphs, we use \textit{networkx} \cite{SciPyProceedings_11}. To validate disassembly directions, we use \textit{coal} \cite{coalweb} in combination with \textit{coacd} \cite{wei2022approximate} for collision detection.

\subsection{Semi-automatic learning from relational assembly models}\label{subsec:semi_automatic_learning}
The main problem with all geometric-symbolic approaches is that they mainly suffer with the underlying data quality. It is extremely important that the information model, which is normally obtained from CAD, is based on defined semantic rules and geometric tolerances. This is normally not the case in practice. To solve this issue, we provide a solution which can learn the basic attributes from geometric data. As an underlying information model, we build on the commonly used relational assembly model suggested in \cite{Mosemann2000}. For this, we learn a mapping $ f $ \eqref{eq:1}, where $ \mathcal{FG} \in \{plane, line, point, circle \} $ describes the geometric element on a component $ C_i $ and $ \mathcal{SSR} \in \{ congruent, concentric, screwed \}$ the symbolic spatial relation between two connected component faces.
\begin{equation} \label{eq:1}
    f: \mathcal{FG} \mapsto \mathcal{SSR}.
\end{equation}

To create a generalized mapping $ f $ we integrate PointNet++ \cite{Qi2017} and a clustering approach. 

\textbf{\textit{Learning relational assembly models.}} For this purpose, the CAD data is meshed, and the resulting mesh is discretized. We combine an equally distributed sampling with focused sampling over sharp edges to obtain point clouds which preserve the finer geometric features like holes of the underlying CAD data. Hereby, every $ C_i $ is presented  as a set of points $P^{(i)}_\mathcal{C}$, where each point is a vector of its coordinate $(x,y,z)$, face normal $(n_{x},n_{y},n_{z})$ and a binary component label $ (l)$. Coordinates are normalized within the range of $ [-1, 1] $. We adapt the PointNet++ semantic segmentation network for the input of $\mathbb{R}^{n\times 7}$ and use the stacked point clouds of two components $(P^{(i)}_\mathcal{C}, P^{(j)}_\mathcal{C})$ for a pairwise relation prediction. Two approaches are used to extract valid relations, one for the $congruent$ and one for the $concentric$ and $screwed$ relations. \autoref{fig:figure_2_filtering} shows the approaches using a simplified example. To gather valid $congruent$ relations, the initial step is to cluster the points by similar normals. These clusters are further filtered by mapping back points to the mesh, so that every group of points lies on a single planar facet. Subsequently, pairs of groups with opposing normals are evaluated for intersection of their planar face projection. For valid $concentric$ and $screwed$ relations, the steps of clustering normals and mapping back is equivalent, only with a greater angle deviation and cylindrical faces. Lastly, the groups of points are evaluated by alignment of their axes as well as intersection of their oriented bounding boxes.

\begin{figure*}[!ht]
    \centering
    \includegraphics[width=\textwidth]{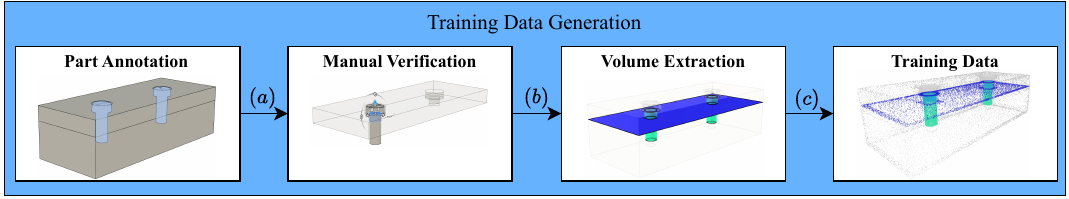}
    \caption{Procedure for the generation of training data. First, threaded components are identified. The automatically identified $ \mathcal{SSR} $ (a) can be manually verified. Areas of faces with $\mathcal{SSR}$ are thickened to create relation volumes (b). Finally, points are sampled from the mesh surface, where points within the relation volumes get labeled (c). \sqboxblack{color_congruent} stands for $congruent$, \sqboxblack{color_concentric} for $concentric$, and \sqboxblack{color_screwed} for $screwed$ relations.}
    \label{fig:figure_3_data_generation}
\end{figure*}

\textbf{\textit{Training data generation.}} To generate the necessary data for training, we use a semi-automated approach that combines \textit{Human-in-the-loop} annotation with automatic geometric analysis of basic topological and geometrical entities. Initially, threaded components are to be annotated. Afterwards, relations are generated by a geometric analysis of the basic CAD data. Given the assembly group $ \mathcal{AG} = \{C_1, C_2, ..., C_n\} $ and geometric data \textit{geoData} that contains their meshes, we gather component pairs $(C_i, C_j)$, where $i\neq j$, with potential $ \mathcal{SSR} $ by checking for intersections within a given tolerance. Matching sets of planar or cylindrical faces may induce a $ congruent $ or $ concentric $ relation if their parameters satisfy the given conditions. For planar faces, these conditions are parallelism, coincidence, and overlap, whereas cylindrical faces must be concentric and overlapping. Given the annotation of threaded components, $ concentric $ and $ screwed $ relations can be distinguished. In the last step, the user confirms the found relations and can add manual relations. This ensures the suggested $ \mathcal{SSR} $ of even suboptimal modeled CAD data with varying semantic rules and geometric tolerances. We determine the actual region on each face that has a relation on $ C_i $ by intersection with it's corresponding face on $ C_j $. By enlarging these intersections we create a volume, in which each point that lies within is respectively labeled. The procedure is shown in \autoref{fig:figure_3_data_generation}.

\subsection{Geometric-symbolic assembly planning}
\label{subsec:geometric_symbolic_planner}

The geometric-symbolic assembly module is based on the ideas from \cite{Friedrich2018}. We extend it with a decay factor to improve runtime and reformulate the principle of the disassembly space so that it is generally applicable for the relational assembly model.

Due to the NP-hardness of the planning problem, we do not consider the possible degrees of freedom ($DoF$) of each component to all the others in order to remove them. The resulting combinatorics are responsible for the fact that the problem cannot be calculated efficiently, particularly for large assemblies. Instead, we propose a two-stage approach. First, we compute the relative $DoF$ from connected components, calling this relative assembly directions. Second, the relative directions are transformed into absolute assembly directions based on visibility and collision detection, applying a divide-and-conquer strategy. For this, we extend the formulation to the occurring relative disassembly space $ \mathbb{W}_{D,r} $ and the absolute ones $ \mathbb{W}_{D,a} $, whereby $ \mathbb{W}_{D,r} \subseteq \mathbb{W}_{D,a} $. The spaces are equivalent in the scenario where all components are visible and there is no collision between them, as we will demonstrate later.

\theoremstyle{definition}
\newtheorem{definition}{Definition}[section]
\begin{definition}[Relative disassembly space $ \mathbb{W}_{D,r} $]
Consists of all $ \mathbb{W}_{D,r}^{(i)} \in SE(3) $ that describes the degrees of freedom $ d_{r}^{(i)} \in \mathbb{R}^6 $ that a component $ C_i $ can perform in its current assembly state, relative to $ C_j \in \mathcal{AG}, \forall j  $.
\end{definition}

\begin{definition}[Absolute disassembly space $ \mathbb{W}_{D,a} $]
Consists of all $ \mathbb{W}_{D,a}^{(i)} \in SE(3) $ that describes the degrees of freedom $ d_{a}^{(i)} \in \mathbb{R}^6 $ that a component $ C_i $ can perform in its current assembly state, to disassemble it without a collision from $ C_j \in \mathcal{AG},  \forall j $.
\end{definition}
\noindent
Through this definition, we have the possibility to easily compute the relative disassembly directions and build a mapping $ f_{ra}: d_{r}^{(i)} \mapsto d_{a}^{(i)} $, based on visibility and collision detection for the absolute disassembly directions, without considering this as a combinatorial problem.

\begin{figure*} [!ht]
    \centering
    \subfloat[]{\includegraphics[width=.24\textwidth]{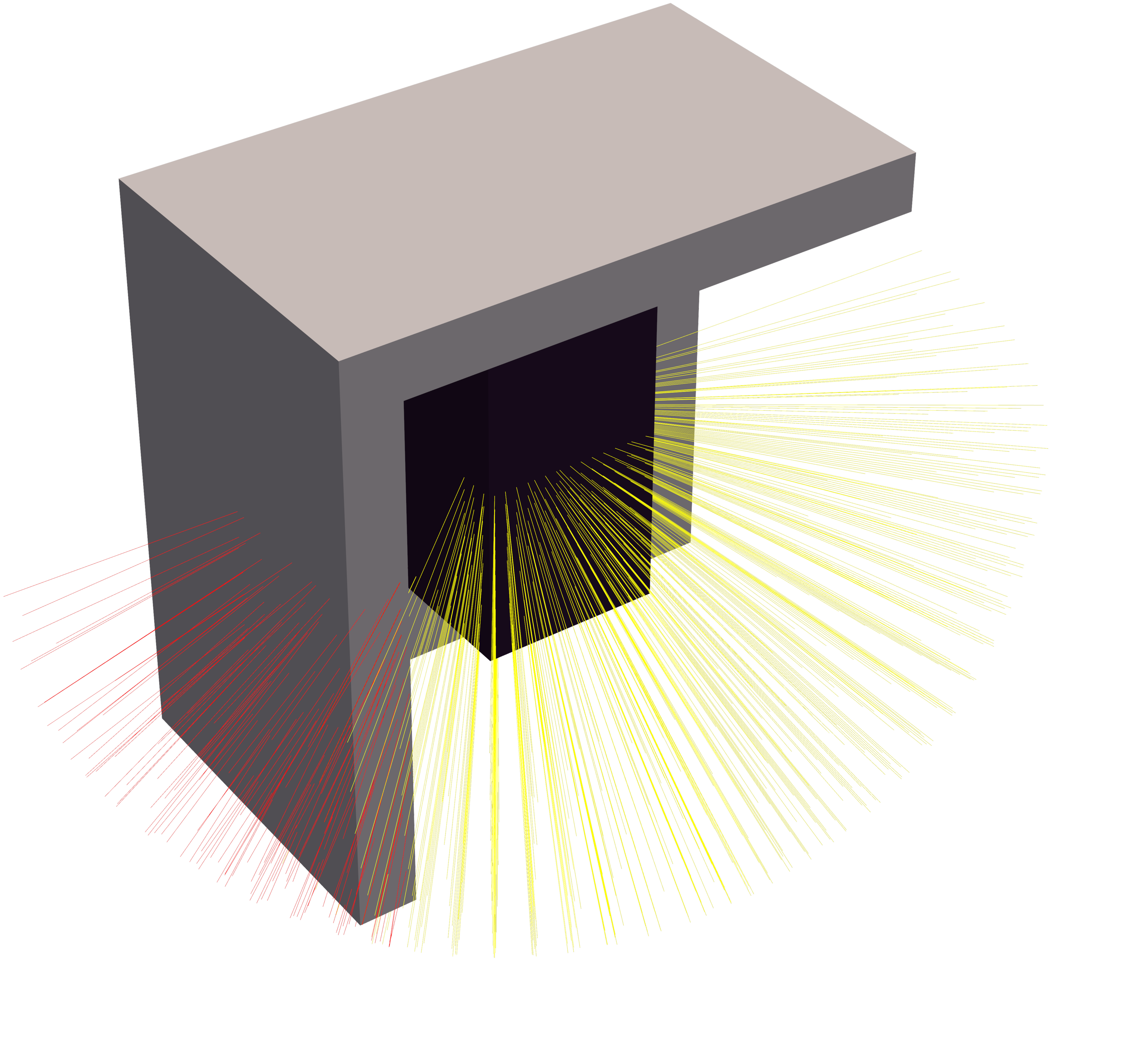}%
    \label{figure_4a_raycast}}
    \hfil
    \subfloat[]{\includegraphics[width=.24\textwidth]{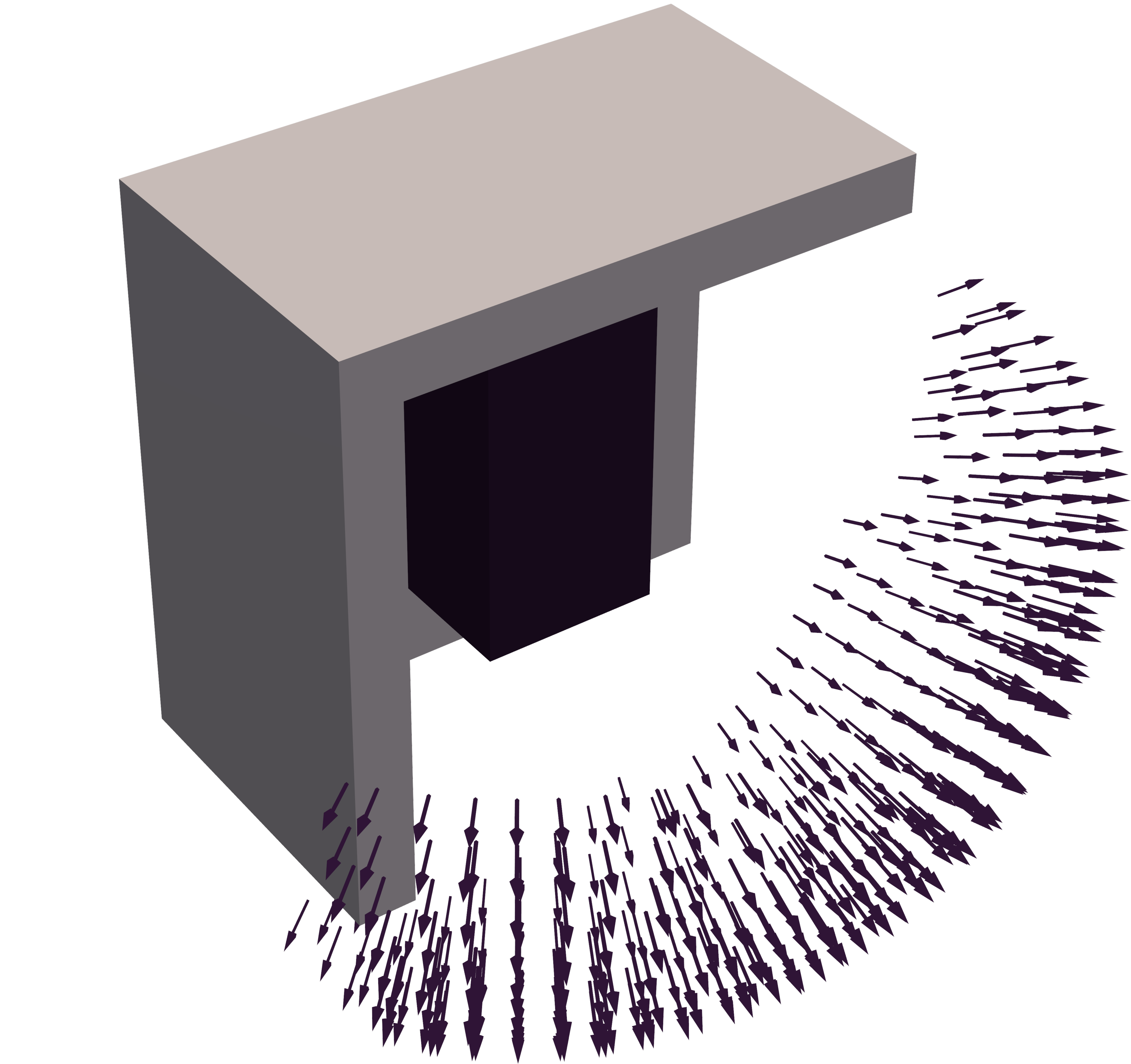}%
    \label{fig:figure_4b_raycast_d_r}}
    \hfil
    \subfloat[]{\includegraphics[width=.24\textwidth]{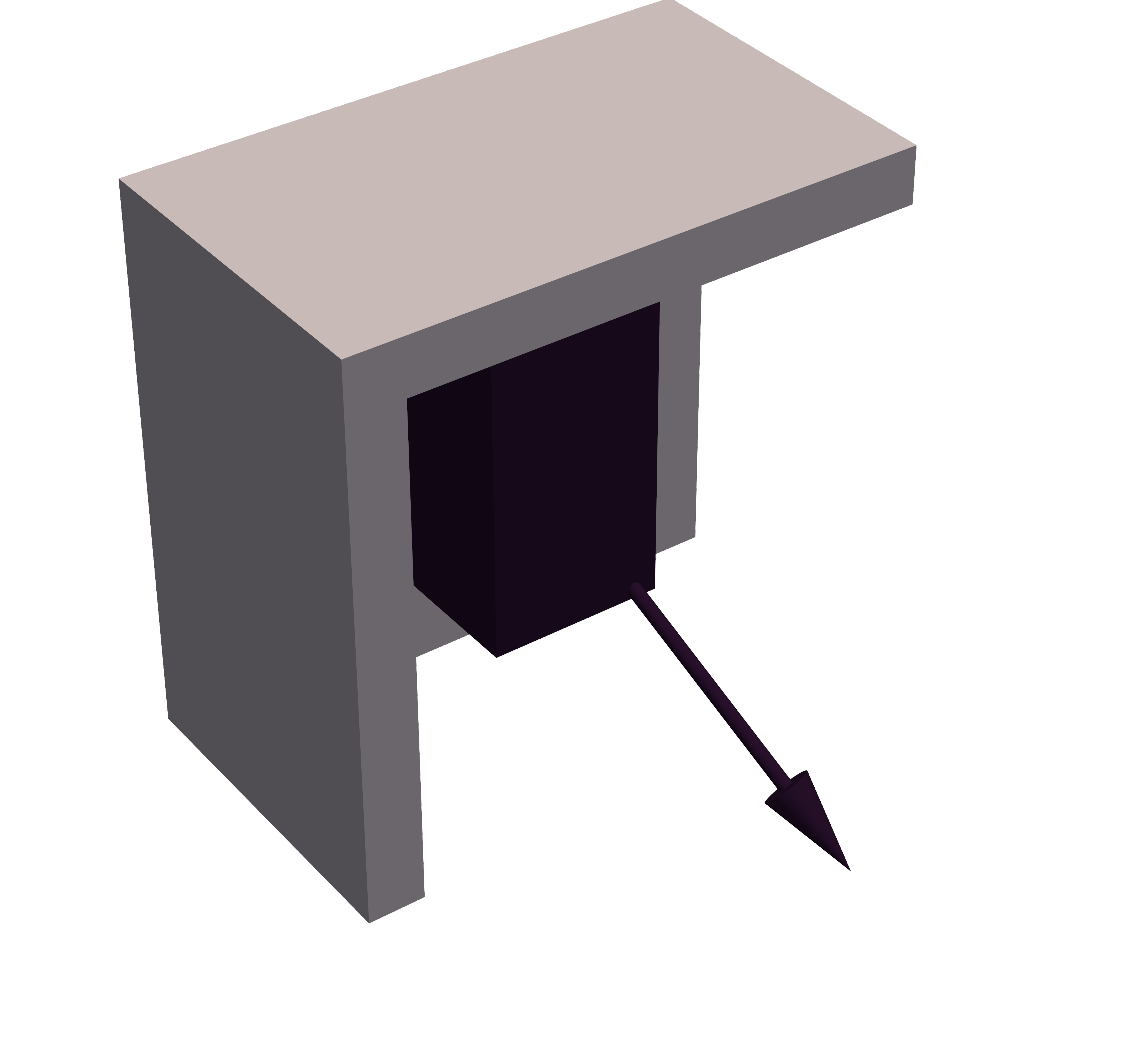}%
    \label{fig:figure_4c_dis_dir}}
    \caption{Example of the mapping $ f_{ra}: d_{r}^{(i)} \mapsto d_{a}^{(i)} $, without considering ASP as a combinatorial problem. (a) Relative Disassembly Space $\mathbb{W}_{D,r}^{(i)} $, the directions are used to cast rays. Rays hitting the component are yellow, occluded rays red. (b) Part of the disassembly space $ d_{r}^{(i)} $ to consider for disassembly. (c) disassembly direction $ d_{a}^{(i)} $ resulting from clustering and accumulation.}
    \label{fig:figure_4_raycast}
\end{figure*}

\RestyleAlgo{ruled}
\SetKwComment{Comment}{/* }{ */}
\begin{algorithm}[!ht]
\caption{Relative assembly directions}\label{alg:one}
\KwData{$ G_{\mathcal{SSR}}(C, \mathcal{SSR}) , geoData. $}
\KwResult{$ G_{sdof}(C,sdof) , d_{r}^{(i)}. $}
\While{$i \leq geoData.length $}{
    $ \mathbb{W}_{D,r,init}^{(i)} \gets $ sample($geoData.mesh(i)) $\\
    $ \mathbb{W}_{D,r,proj}^{(i)} \gets $ project($ \mathcal{SSR}(i) $)\\
    $ \mathbb{W}_{D,r}^{(i)} \gets $ intersect($ \mathbb{W}_{D,r,proj}^{(i)} $)\\
    $ [d_{r}^{(i)}, sdof(i)] \gets $ assignment($ \mathbb{W}_{D,r}^{(i)} )$\\
    $i \gets i+1$
    }
\end{algorithm}

\begin{figure} [!ht]
    \centering
    \includegraphics[width=0.47\textwidth]{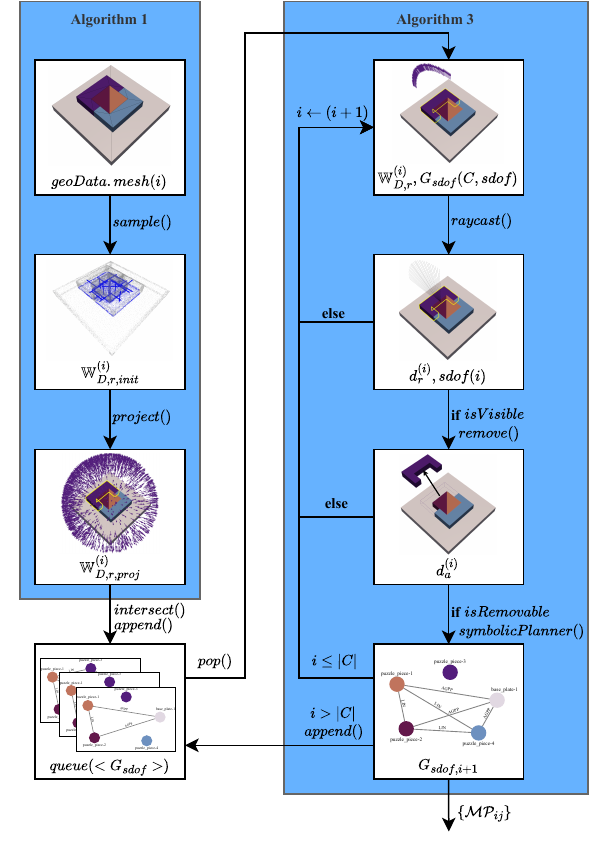}
    \caption{Simplified flowchart of the geometric-symbolic disassembly planning process. \autoref{alg:one} computes the initial $G_{sdof}$ and $d_r$. For each graph state, \autoref{alg:three} executes the geometric-symbolic planner, generating the set of motion primitives $\{\mathcal{MP}_{ij}\}$ for all components $C_i$ that are removable within the current visibility space $j$, while updating the graph to the next state $G_{sdof,i+1}$ after each successful component removal.
    }
    \label{fig:figure_5_flowchart}
\end{figure}

\textbf{\textit{Computation of the relative disassembly directions.}} 
We uniformly sample surface points from all components $C$. Their corresponding normal vectors form an approximately enclosed distribution on the unit sphere. This initial space $ \mathbb{W}_{D,r,init}^{(i)} $ is used to project the $ \mathcal{SSR} $ on it. The resulting $ \mathbb{W}_{D,r,proj}^{(i)} $ are intersected to compute the final relative disassembly space $ \mathbb{W}_{D,r}^{(i)} $ for each component $ i $. This enables the assignment from $ sdof \in\{ fix,lin, rot, fits, agpp, free \}$ to describe the symbolic state of the degree of freedom, compare \cite{Mosemann2000}, as well as the relative disassembly directions $ d_{r}^{(i)} $. 
The determination of the relative translational disassembly space is reduced to sorting the interval boundaries followed by a linear comparison of the resulting ordered lists. Since sorting requires $\mathcal{O}(n \log n)$ and $\mathcal{O}(m \log m)$ for lists of lengths $n$ and $m$, respectively, and the subsequent comparison can be performed in $\mathcal{O}(n + m)$. This leads to a overall complexity, which is dominated by the sorting step, where $n$ and $m$ represent the samples of the translational disassembly directions. Hence, the relative disassembly space of a component can be computed in $\mathcal{O}(n \log n)$ for $n \geq m$.
Building on this, we obtain the corresponding graph $ G_{sdof}(C,sdof) $, which is the foundation for symbolic planning. The procedure is explained in \autoref{alg:one} and in \autoref{fig:figure_5_flowchart} in the left section.

\textbf{\textit{Computation of absolute disassembly directions and symbolic planning.}} Based on $ G_{sdof}(C,sdof) $ the absolute disassembly directions are computed. For that, the algorithm starts with the component with the lowest connection costs. Here, the connection costs $ \mathcal{CC} $ are higher the greater the constraint imposed by the respective relations, defined in \eqref{eq:constraints2}.

\RestyleAlgo{ruled}
\begin{algorithm}[!ht]
\caption{Hierarchical Disassembly Planning}\label{alg:two}
\KwData{$ G_{sdof}(C,sdof) , d_{r}^{(i)}, geoData. $}
\KwResult{$ \mathcal{MP}_{ij}, d_{a}^{(i)}. $}
$queue$.append($G_{sdof}$) \\
\While{queue \text{is not empty}}
{
    $G_{sdof, k} \gets queue$.pop()\\
    $\langle C \rangle \gets $ sortConnectCosts($ G_{sdof, k}(C,sdof)) $\\
    $ j \gets 0 $ \Comment{stateVisibilitySpace}
    \ForEach{$C_i $ in $ \langle C \rangle $}{
        $ { G_{sdof, k + 1}, \mathcal{MP}_{ij}, d_{a}^{(i)} } \gets $ GeometricSymbolicPlanner($ G_{sdof, k}(C,sdof) , geoData $) \\
    }
    $ j \gets j+1 $\\
    $\langle G_{sdof}\rangle \gets $ connectedSubgraphs($G_{sdof, k + 1}$)\\
    \ForEach{$G_{sdof,k}$ in $\langle G_{sdof, s}\rangle  $}{
    $queue$.append($G_{sdof, k})$
    }
}
\end{algorithm}

\RestyleAlgo{ruled}
\begin{algorithm}[!ht]
\SetKwComment{Comment}{/* }{ */}
\SetKwProg{Fn}{Function}{}{end}
\SetKwFunction{FRecurs}{GeometricSymbolicPlanner}%
\caption{Geometric-Symbolic Planning}\label{alg:three}
\Fn{\FRecurs{}}
{
    \KwData{$ G_{sdof, k}(C,sdof) , geoData $}
    \KwResult{$ G_{sdof, k + 1}, \mathcal{MP}_{i}, d_{a}^{(i)} $}
    $\langle C \rangle \gets $ sortConnectCosts($ G_{sdof, k}(C,sdof)) $\\
    $ [isVisible, d_{r}^{(i)}] \gets $ raycast($ C_i, geoData, sdof(i)) $\\
    \If{$isVisible$}{
        $ [d_{a}^{(i)}, isRemovable] \gets $ remove($ C_i, geoData, d_{r}^{(i)}) $\\
          \If{$isRemovable$}{
          $\mathcal{MP}_{i} \gets $ symbolicPlanner($ G_{sdof, i}) $\\
        }
    }
    $G_{sdof, k + 1} \gets$ symbolicPlanner($\mathcal{MP}_{0}$)\\
}
\end{algorithm}

\begin{equation} \label{eq:constraints2}
    \mathcal{CC}_{free} < \mathcal{CC}_{agpp} < \mathcal{CC}_{fits} < \mathcal{CC}_{rot} < \mathcal{CC}_{lin} < \mathcal{CC}_{fix}.
\end{equation}

 For the component with the lowest connection costs, an absolute disassembly direction is derived from the relative disassembly directions. For this, ray-casting is performed along the relative disassembly directions $d_r^{(i)}$. All rays that reach the component without being occluded are clustered and accumulated to obtain a set of possible disassembly directions. These directions are evaluated sequentially using a collision-based removal test. As soon as a collision-free removal is identified, the corresponding direction is accepted as a valid absolute disassembly direction $d_a^{(i)}$ and forwarded to the symbolic planner. The procedure is illustrated in \autoref{fig:figure_4_raycast}.

 This procedure enables us to plan a set of manipulation primitives $ \mathcal{MP}_{ij} = \{\{\mathcal{MP}_{00}, ..., \mathcal{MP}_{0n}\} , ..., \{\mathcal{MP}_{m0}, ..., \mathcal{MP}_{mn}\}\} $ with symbolic actions which fulfill the task. They are ordered according to the resulting $ n$-visibility spaces in which a component falls for the current assembly state. The symbolic planner is adopted from \cite{Friedrich2018}. It uses a symbolic disassembly logic, which describes the state transitions $ \delta: sdof_{k}(C) \times \mathcal{MP} \mapsto sdof_{k+1}(C) $. Here, a manipulation primitive is a symbolic action $ \mathcal{MP} \in \{move, twist, pull, put\} $, which describes the robot capabilities. With $\delta$ to obtain the next graph $ G_{sdof, k+1}$ by removing the cheapest component. In case of the creation of disconnected subgraphs, we view each one as a single connected graph and repeat this process, until every component of every graph is disassembled. \autoref{alg:two}, \autoref{alg:three} and the right section of \autoref{fig:figure_5_flowchart} summarize our approach. The pure planning approach, based on the relational graph, results in a running time with $ \mathcal{O}(n) $, where $n = m_i\cdot|C|$, $m_i \ll |C|$. Here, $m_i$ are the state transitions for the number of components $|C|$ in the overall assembly. The underlying collision detection based on bounding volume hierarchies results in an average running time in $ \mathcal{O}(n) $ with $n$ tested triangles \cite{klein2005expected}.

\textbf{{Sequence optimization}}. To optimize the valid sequence from the planning for robot execution, we formulate it as a traveling salesman problem. For that, we consider the motion $ t_{m,i} $ and the tool changing cost $ t_{t,j} $ between different manipulation primitives $ \mathcal{MP}_k $ and $ \mathcal{MP}_{k+1} $ in terms of time. We use an integer linear programming (ILP) approach for small instances ($ |\mathcal{MP}| < 20 $) and a Nearest-Neighbor (NN) algorithm for larger ones to find $t_s^*$ \eqref{eq:2}.
\begin{equation} \label{eq:2}
    t_s^* = \mathop{\arg \min}\limits_{t_m, t_t  \in T} ( \sum_{i=0}^N t_{m,i} + \sum_{j=0}^M t_{t,j}).
\end{equation}
ILP guarantees an optimal solution with respect to costs opposite to NN for our local sequence, but runs in $ \mathcal{O} (| \mathcal{MP} |!) $ compared to $ \mathcal{O} (| \mathcal{MP} |^2) $. The ILP Solution is used as a baseline for the optimality of the sub-spaces.

\section{Experimental Results}

\noindent For a comprehensive validation of the approach, we evaluate the preprocessing and the planner. The preprocessing is evaluated in \autoref{subsec:eval_preprocessing} using a test dataset with varying tolerances with assemblies from our dataset presented in  \autoref{subsec:dataset}, consisting of our models as well as known models from literature. For the evaluation of our planner in \autoref{subsec:eval_planner} we use our dataset as well as the test dataset used in ASAP\cite{Tian2024} to allow a better comparison of the planning capabilities. We perform our experiments on a workstation featuring an AMD Ryzen 7 5700X processor, 64 GB of RAM, and a NVIDIA GeForce RTX 4090 GPU.

\subsection{Assembly dataset}
\label{subsec:dataset}
\begin{figure*}[!ht]
    \centering
    \includegraphics[width=0.85\textwidth]{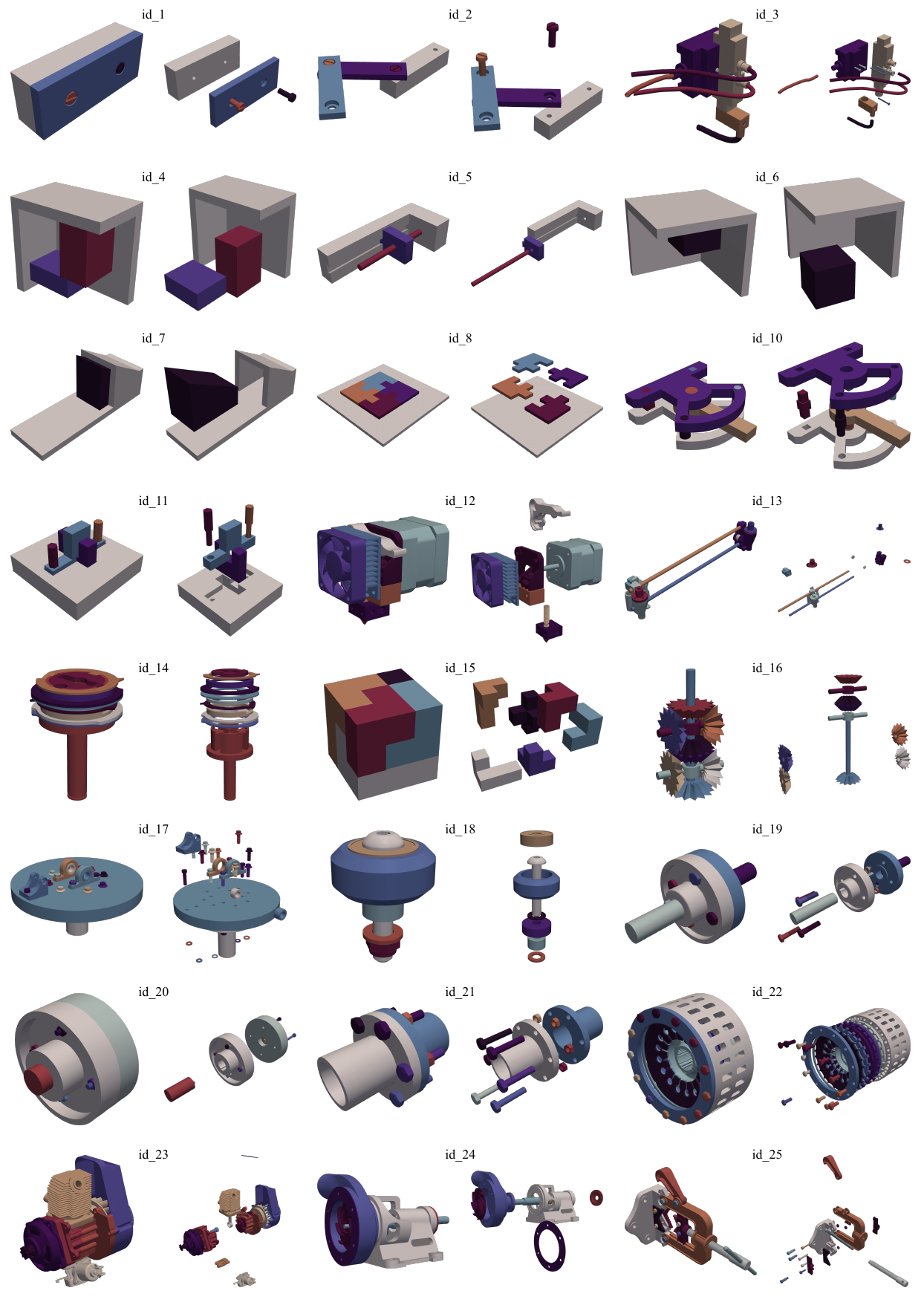}
    \caption{Examples from the used dataset. The assemblies are shown in an assembled and disassembled state. Sources can be found in \autoref{tab:dataset_sources}.}
    \label{fig:figure_6_assembly_dataset}
\end{figure*}

Our dataset contains $25$ CAD assemblies, ranging from $2$ to $45$ components, and from simple geometries to more complex real world assemblies. \autoref{fig:figure_6_assembly_dataset} illustrates samples of the dataset, showing each assembly in both its assembled and disassembled state. The set contains models from \cite{AutoMate, lupinettiContentbasedMulticriteriaSimilarity2019, willis2021joinable, luoFMBFunctionalManipulation2024, GrabCADMakingAdditive, CranfieldBenchmark} as well as own models. Sources can be found in \autoref{tab:dataset_sources}. Minor modifications were required to a small number of models to ensure that they could be physically disassembled. All of these can be disassembled with translational movements and do not require subassembly identification. Due to licensing restrictions, we can not redistribute the assemblies, but the used models and their source are listed in the appendix in \autoref{tab:dataset_sources}.

\begin{table*}[!ht]
\centering
\caption{Evaluation metrics for the semantic feature learning network and extracted assembly relations on the complete test set and manually selected tolerance-specific subsets.}
\resizebox{\textwidth}{!}{%
\begin{tabular}{@{}lcccccccccc@{}}
\toprule
                      & \multicolumn{2}{c}{\textbf{accuracy(\%)}} & \multicolumn{5}{c}{\textbf{mIoU(\%)}}                     & \multicolumn{3}{c}{\textbf{accuracy of extracted relations(\%)}} \\ \cmidrule(l){2-3} \cmidrule(l){4-8} \cmidrule(l){9-11} 
\multicolumn{1}{c}{}  & avg. class            & overall           & average & $none$ & $congruent$ & $concentric$ & {$screwed$} & $congruent$          & $concentric$          & $screwed$          \\ \midrule
\textbf{all}          & 72.6                  & 90.3              & 54.6    & 90.7   & 54.9        & 36.0         & 36.6      & 83.19                & 62.2                  & 47.06             \\
\textbf{no tolerance} & 76.2                  & 90.5              & 55      & 91.4   & 64.0        & 29.94        & 35.3      & 100                  & 62.5                  & 60                \\
\textbf{f}            & 76.6                  & 91.2              & 56.0    & 92.2   & 63.8        & 34.0         & 34.0      & 84.62                & 62.5                  & 40                \\
\textbf{m}            & 77                    & 91                & 55.4    & 92.0   & 62.6        & 34.5         & 32.6      & 81.48                & 64.52                 & 45                \\
\textbf{c}            & 76.6                  & 91.2              & 56      & 92.2   & 63.8        & 34.0         & 34.0      & 84.62                & 62.5                  & 55                \\
\textbf{v}            & 73.2                  & 89.8              & 52.7    & 90.8   & 57.7        & 27.2         & 35.0      & 78.57                & 59.38                 & 45                \\ \bottomrule
\end{tabular}%
}
\label{tab:network_extraction_metric}
\end{table*}

\begin{table}[!ht]
\centering
\caption{Planning Success with Extracted Relations from Semantic Feature Learning without manual Annotation.}
\begin{tabular}{@{}ccccccc@{}}
\toprule
\multicolumn{1}{l}{} & \textbf{overall} & \textbf{no tolerance} & \textbf{f} & \textbf{m} & \textbf{c} & \textbf{v} \\ \midrule
\textbf{Success}     & 146              & 10                    & 33         & 34         & 35         & 34         \\
\textbf{Failed}      & 58               & 2                     & 15         & 14         & 13         & 14         \\ \bottomrule
\end{tabular}
\label{tab:rel_extraction_succ}
\end{table}

\subsection{Evaluation of preprocessing}
\label{subsec:eval_preprocessing}
In the first step, we evaluate the results of our method proposed in \autoref{subsec:semi_automatic_learning} for semi-automatic learning of a fully-defined relational assembly model. For this purpose we use a subset of our Assembly Dataset that consists of well defined CAD data to enable automatic generation of the ground truth annotations (\textit{id\_1}-\textit{id\_10} and \textit{id\_19}-\textit{id\_21}). To augment our dataset, component pairs are combined with additional parts while preserving the corresponding relations. To evaluate the robustness of the semantic feature learning network with respect to imperfect CAD data, we introduce four tolerance classes inspired by ISO 2768-1 \cite{ISO2768_1}, namely \textit{f}, \textit{m}, \textit{c}, and \textit{v}. For every component pair, each component is translated independently in the positive and negative direction of its relation by the upper bound of the respective tolerance class. This results in 16 additional samples per base pair. The characteristic length $l$ used to determine the applicable tolerance for a component $C_i$ is computed as the diagonal length of its axis aligned bounding box.

The resulting dataset contains 4198 component pairs, which are divided into 3358 training samples, 420 validation samples, and 420 test samples. From the test set, 12 representative component pairs are manually selected for the tolerance-specific evaluation, with each relation type represented five times.

The semantic feature learning network is first evaluated using the point-wise metrics reported in \autoref{tab:network_extraction_metric}. Besides the complete test set (\textit{all}), the manually selected tolerance-specific test subset is reported separately. On the complete test set (\textit{all}), the network achieves an overall point accuracy of 90.3\% and an average class accuracy of 72.6\%. Since the \textit{none} class dominates the dataset, the mean Intersection over Union (mIoU) provides a more representative measure of segmentation quality than the overall accuracy. The network achieves an average mIoU of 54.6\%, with noticeable differences between the individual relation classes. \textit{Congruent} relations achieve the highest IoU (54.9\%), whereas \textit{concentric} and \textit{screwed} relations achieve 36.0\% and 36.6\%, respectively. The lower performance of the latter two classes is likely caused by their lower representation in the dataset as well as their similar geometric appearance. Since threads are typically not modeled explicitly in CAD assemblies, \textit{screwed} connections are often represented solely by intersecting cylindrical surfaces, which reduces the distinguishability between both relation types. Inspection of the predicted relations further confirms that \textit{concentric} and \textit{screwed} connections are frequently confused with each other.

The influence of geometric tolerances on the learned semantic features remains comparatively small. For the tolerance classes \textit{f}, \textit{m}, and \textit{c}, both the overall accuracy and the average mIoU remain nearly constant, with average mIoU values between 55.4\% and 56.0\%. Only the largest tolerance class \textit{v} causes a noticeable decrease in performance, reducing the average mIoU to 52.7\% and the overall accuracy to 89.8\%. These results indicate that the proposed semantic feature learning approach generalizes well to moderate geometric deviations while remaining robust under realistic manufacturing tolerances.

To evaluate the complete preprocessing pipeline, the predicted point labels are subsequently processed by the proposed relation extraction algorithm. The relations extracted using the method presented in \autoref{subsec:semi_automatic_learning} are compared against geometrically extracted and manually verified CAD relations, which serve as the ground truth. A relation is considered correct if both its origin and direction deviate from the reference by less than predefined translational and angular thresholds. The relation extraction accuracies are summarized in \autoref{tab:network_extraction_metric}. Despite the moderate point-wise segmentation performance, the geometric post-processing reconstructs $congruent$ relations with an accuracy of 83.2\%, demonstrating that the extraction stage successfully compensates for many local segmentation errors. The accuracies for $concentric$ (62.2\%) and $screwed$ (47.1\%) relations remain lower, reflecting the difficulties already observed during semantic segmentation. Furthermore, only minor variations can be observed across the different tolerance classes, indicating that the proposed extraction algorithm is largely insensitive to moderate geometric deviations.

Finally, the extracted relations are evaluated within the complete geometric-symbolic planning pipeline. As shown in \autoref{tab:rel_extraction_succ}, 146 of 204 test assemblies are successfully planned, corresponding to an overall success rate of 71.6\%. More importantly, the planning success remains nearly constant across all tolerance classes, varying only between 68.8\% and 72.9\%. This indicates that the introduced tolerances have only a limited impact on the overall planning performance. Instead, the remaining failures are primarily associated with inaccuracies in the predicted relations, particularly the confusion between $concentric$ and $screwed$ relations.

\subsection{Evaluation of the planner}
\label{subsec:eval_planner}
In the second evaluation, we assess the geometric-symbolic planner proposed in \autoref{subsec:geometric_symbolic_planner} with respect to runtime and sequence optimization using our dataset introduced in \autoref{subsec:dataset}. We further use ASAP as a baseline to compare our planner in terms of success and planning time on the ASAP test dataset. 

\begin{figure*}[!ht]
    \centering
    \subfloat[]{\includegraphics[width=.49\textwidth]{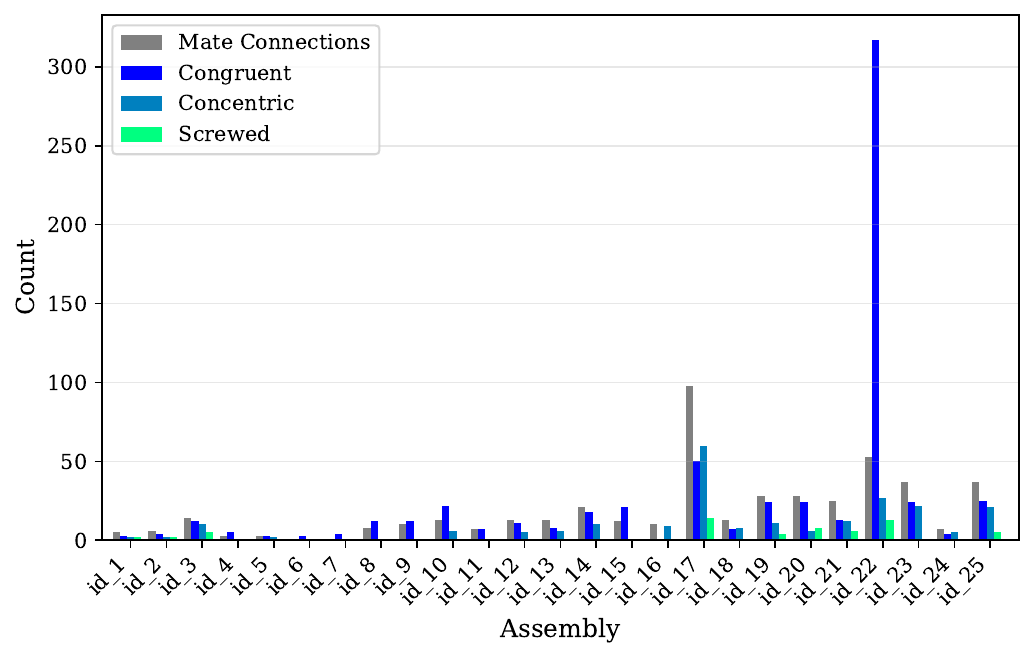}}
    \label{fig:figure_7a_mate_connections}
    \hfill
    \subfloat[]{\includegraphics[width=.49\textwidth]{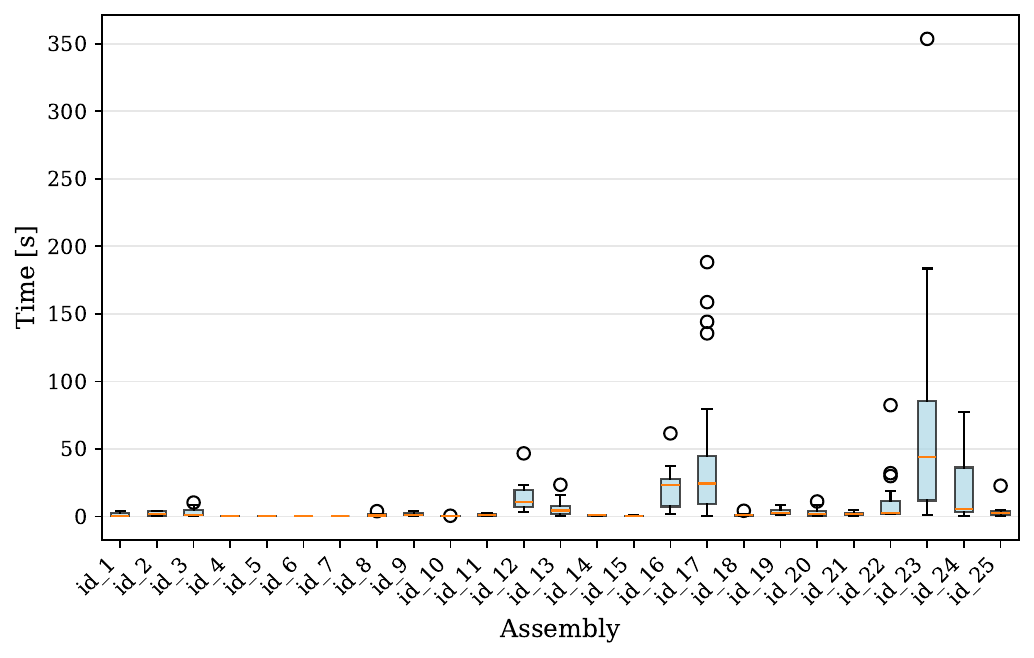}}
    \label{fig:figure_7b_planning_times}
    \caption{Result of the planner over the proposed dataset. (a) shows the number of connections in the respective assemblies. A \textit{Mate Connection} stands for any amount of existing relations $congruent$, $concentric$ or $screwed$ between two components $C_i$ and $C_j$. (b) shows the distribution of the planning times per component by assembly} 
    \label{fig:figure_7_planner_metrics}
\end{figure*}

\textbf{\textit{Geometric disassembly and symbolic sequence planner.}}\label{subsubsec:eval_planner} \autoref{fig:figure_7_planner_metrics} shows the distribution of the relation count (a) and the average planning time per removable component (b). The reported planning times are averaged over 10 independent planning runs for each assembly. The relation histogram demonstrates that the evaluated assemblies cover a broad range of geometric complexity, ranging from only three to 357 geometric relations. Despite this diversity, the boxplot indicates that most planning queries are completed within only a few seconds. Only a small number of components exhibit substantially longer planning times, which predominantly occur in assemblies containing geometrically complex parts rather than their count of relations. The left part of  \autoref{tab:planner_metrics} breaks down the runtime into the mean total planning time $\bar{t}_{\mathcal{AG}}$ and the mean execution time of the major stages of the proposed planning pipeline $\bar{t}_{d_r}$, $\bar{t}_{d_a}$, $\bar{t}_{\mathcal{C}}$ and $\bar{t}_{sdl}$. $\bar{t}_{\mathcal{AG}}$ represents the mean end-to-end planning time for the complete assembly. The remaining columns report the mean execution time of the corresponding planning stages, namely ray-casting ($\bar{t}_{d_r}$), candidate direction generation ($\bar{t}_{d_a}$), collision checking ($\bar{t}_{\mathcal{C}}$), and symbolic disassembly logic ($\bar{t}_{sdl}$). Note that the reported stage runtimes do not constitute a partition of the total planning time. During planning, individual stages can be invoked multiple times for a single component.

\begin{table*}[!ht]
\centering
\caption{Metrics of the Planner and Optimization. 
Planning times are averaged over 10 independent runs per assembly. For each assembly, the longest planning stage is highlighted in bold. Individual planning stages may be executed multiple times and therefore do not sum to the total planning time.}
\begin{tabular}{@{}cccccccccccc@{}}
\toprule
\textbf{}         &            &           & \multicolumn{5}{c}{\textbf{Geometric-Symbolic Planning $[s]$}}                                                     & \multicolumn{4}{c}{\textbf{Sequence Optimization}}        \\ \cmidrule(l){4-8} \cmidrule(l){9-12} 
\textbf{Assembly} & $n_{comp}$ & $n_{rel}$ & $\bar{t}_{\mathcal{AG}}$ & $\bar{t}_{d_{r}}$ & $\bar{t}_{d_{a}}$ & $\bar{t}_{\mathcal{C}}$ & $\bar{t}_{sdl}$ & $\bar{t}_{deg, c}$ & $Cost_{NN}$ & $Cost_{ILP}$ & $t_{imp,s}\ [\%]$ \\ \midrule
id\_1             & 4          & 7         & 4.52                     & 0.05              & \textbf{3.80}     & 0.36                    & 0.31            & 16.30              & 20.05       & 20.05        & 0.00           \\
id\_2             & 5          & 8         & 8.92                     & 0.08              & \textbf{7.30}     & 0.86                    & 0.66            & 11.27              & 20.11       & 20.11        & 0.00           \\
id\_3             & 12         & 27        & 34.20                    & 0.19              & 0.09              & \textbf{33.73}          & 0.15            & 589.68             & 30.59       & 30.58        & 0.00           \\
id\_4             & 3          & 5         & 0.17                     & 0.01              & \textbf{0.11}     & 0.03                    & 0.01            & 18.46              & 0.05        & 0.05         & 0.00           \\
id\_5             & 3          & 5         & 0.09                     & 0.02              & 0.00              & \textbf{0.06}           & 0.01            & 21.50              & 10.00       & 10.00        & 0.00           \\
id\_6             & 2          & 3         & 0.07                     & 0.01              & \textbf{0.04}     & 0.01                    & 0.00            & -                  & -           & -            & -              \\
id\_7             & 2          & 4         & 0.04                     & 0.01              & 0.00              & \textbf{0.02}           & 0.00            & -                  & -           & -            & -              \\
id\_8             & 5          & 12        & 5.26                     & 0.04              & \textbf{3.82}     & 0.08                    & 1.31            & 11.51              & 0.18        & 0.18         & 0.00           \\
id\_9             & 5          & 12        & 6.64                     & 0.05              & \textbf{3.92}     & 0.06                    & 2.61            & 67.46              & 0.03        & 0.03         & 0.00           \\
id\_10            & 8          & 28        & 1.43                     & 0.03              & 0.01              & 0.61                    & \textbf{0.76}   & 14.98              & 0.48        & 0.42         & 12.5           \\
id\_11            & 5          & 7         & 4.51                     & 0.05              & \textbf{2.82}     & 0.35                    & 1.28            & 17.23              & 0.16        & 0.16         & 0.00           \\
id\_12            & 9          & 17        & 126.57                   & 0.27              & 0.40              & \textbf{111.71}         & 2.63            & 49.84              & 30.13       & 30.13        & 0.00           \\
id\_13            & 12         & 14        & 67.30                    & 0.53              & 6.92              & \textbf{33.11}          & 0.64            & 84.60              & 1.18        & 1.12         & 5.08           \\
id\_14            & 11         & 28        & 6.87                     & 0.09              & 0.01              & \textbf{4.26}           & 2.46            & 18.36              & 40.01       & 40.01        & 0.00           \\
id\_15            & 6          & 21        & 2.14                     & 0.02              & 0.13              & 0.09                    & \textbf{1.89}   & 38.72              & 0.13        & 0.12         & 0.08           \\
id\_16            & 10         & 10        & 206.43                   & 0.13              & 0.01              & \textbf{191.89}         & 0.03            & 78.79              & 0.05        & 0.05         & 0.00           \\
id\_17            & 45         & 124       & 1684.84                  & 8.36              & 12.38             & \textbf{1279.25}        & 69.18           & -                  & 22.88       & -            & -              \\
id\_18            & 8          & 16        & 8.86                     & 0.09              & 0.61              & \textbf{7.90}           & 0.21            & 13.29              & 20.01       & 20.01        & 0.00           \\
id\_19            & 14         & 39        & 45.57                    & 0.57              & 1.32              & \textbf{38.55}          & 5.07            & 175.41             & 20.52       & 20.52        & 0.00           \\
id\_20            & 14         & 38        & 39.99                    & 0.33              & 2.62              & \textbf{29.30}          & 7.68            & 90.76              & 21.16       & 21.10        & 0.00           \\
id\_21            & 14         & 31        & 26.43                    & 0.22              & \textbf{11.33}    & 7.20                    & 7.63            & 313.08             & 21.06       & 21.06        & 0.00           \\
id\_22            & 26         & 357       & 254.64                   & 1.48              & 0.06              & \textbf{246.05}         & 6.24            & -                  & 20.36       & -            & -              \\
id\_23            & 7          & 46        & 1517.90                  & 4.48              & 7.16              & \textbf{1447.13}        & 5.09            & -                  & 0.81        & -            & -              \\
id\_24            & 7          & 9         & 137.34                   & 0.16              & 0.01              & \textbf{55.32}          & 0.02            & 56.16              & 1.14        & 1.14         & 0.00           \\
id\_25            & 20         & 51        & 68.74                    & 0.68              & 6.02              & \textbf{58.97}          & 2.88            & 175.69             & 20.93       & 20.92        & 0.00           \\ \bottomrule
\end{tabular}
\label{tab:planner_metrics}
\end{table*}

\begin{table*}[!ht]
\centering
\caption{Comparison of \frameworkacr{} to baseline with ASAP test dataset.}
\begin{tabular}{ccccccc}
\toprule
                                          &                   & \multicolumn{5}{c}{median runtime for n parts $[s]$}                              \\ \cmidrule(l){3-7} 
Method                                    & Success Rate (\%) & $\leq 5$      & $(5,10]$         & $(10, 20]$     & $(21, 30]$     & $>30$         \\ \midrule
\textbf{\frameworkacr (Ours)}             & \textbf{85.83}             & \textbf{1.32} & \textbf{8.04} & \textbf{15.69} & \textbf{25.35} & \textbf{80.9} \\
ASAP learning (High Budget, 4 Parts Held)\cite{Tian2024} & 82.08    & 43            & 97            & 431            & 1705           & 4190          \\ \bottomrule
\end{tabular}
\label{tab:ASAP_comparison}
\end{table*}

\begin{figure*}[!ht]
    \centering
    \includegraphics[width=0.9\textwidth]{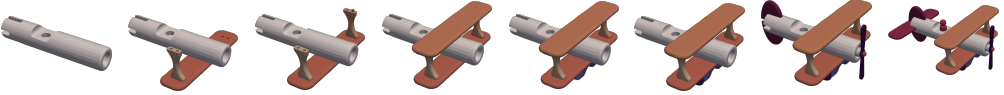}
    \includegraphics[width=0.9\textwidth]{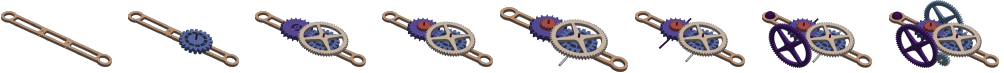}
    \includegraphics[width=0.9\textwidth]{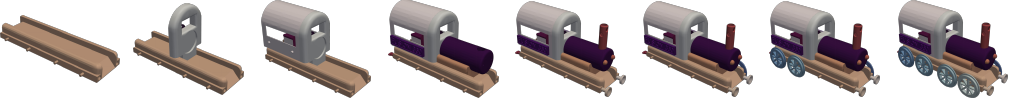}
    \caption{Example assembly sequences of the ASAP test data generated with \frameworkacr.}
    \label{fig:figure_8_assembly_sequences}
\end{figure*}

The runtime breakdown reveals that the computational effort is distributed unevenly across the individual planning stages. The ray-casting stage ($\bar{t}_{d_r}$), which determines the visible subset of the relative disassembly space, contributes only a negligible fraction of the total planning time for all evaluated assemblies. In contrast, the computation of disassembly directions for collision checking ($\bar{t}_{d_a}$) exhibits a higher variability. Since this stage clusters and accumulates the visible subset of the relative disassembly space, its runtime depends on its size and distribution.

The dominant  to the overall planning time is the collision checking stage ($\bar{t}_{\mathcal{C}}$). This behavior is particularly evident for assemblies containing geometrically complex components, resulting in significantly more expensive collision queries. Consequently, the outliers observed in \autoref{fig:figure_7_planner_metrics} (b) are primarily associated with geometric complexity rather than with the number of geometric relations alone. For example, assemblies such as \textit{id\_23} and \textit{id\_24} exhibit substantially longer planning times despite containing only 46 and 9 geometric relations, respectively, illustrating that collision complexity is a stronger predictor of runtime than the relation count.

The symbolic disassembly logic ($\bar{t}_{sdl}$) represents only a comparatively small portion of the total runtime. Its computational effort mainly depends on the assembly topology, as each successful removal requires selecting the corresponding manipulation primitive, updating the symbolic assembly graph, and recomputing the relative disassembly spaces affected by the removed component. Assemblies with many interconnected components or highly connected constraint graphs therefore exhibit moderately increased symbolic planning times, while the overall runtime remains dominated by the geometric processing stages.

Overall, the evaluation demonstrates that the proposed geometric-symbolic planning framework scales to complex industrial assemblies. The results further show that the geometric-symbolic reasoning introduces only a minor computational overhead, while collision checking remains the primary contributor to the overall planning time.

\textbf{\textit{Sequence optimization}} Next, we evaluate the proposed sequence optimization based on the visibility spaces. To enable a quantitative comparison, we define the relative improvement of the optimized sequence cost as

\begin{equation}
    t_{imp,s}=1-\frac{t_{s,ILP}^{*}}{t_{s,NN}^{*}},
\end{equation}

\noindent where $t_{s}^{*}$ denotes the optimal sequence cost obtained by the respective optimization method. Since the nearest-neighbor (NN) heuristic has a computational complexity of $\mathcal{O}(|\mathcal{MP}|^2)$, we additionally evaluate the computational overhead of the ILP optimization using

\begin{equation}
    t_{deg,c}=\frac{t_{c,ILP}}{t_{c,NN}},
\end{equation}

\noindent where $t_c$ denotes the computation time required for sequence optimization.

As summarized in \autoref{tab:planner_metrics}, the ILP optimization achieves the same sequence cost as the nearest-neighbor (NN) heuristic for the vast majority of evaluated assemblies. Only three assemblies exhibit a measurable improvement, with relative sequence cost reductions of 12.5\%, 5.08\%, and 0.08\%, respectively. These results indicate that the proposed NN heuristic already produces near-locally optimal disassembly sequences for most assemblies considered in this work. In contrast, the computational overhead of the ILP optimization is substantial. The runtime degradation factor ranges from approximately 11 to almost 590 for the evaluated assemblies. Overall, the results indicate that the marginal improvements in sequence cost are generally accompanied by a significantly increased computational effort.

\begin{figure*} [!ht]
    \centering
    \subfloat[]{\includegraphics[width=.24\textwidth]{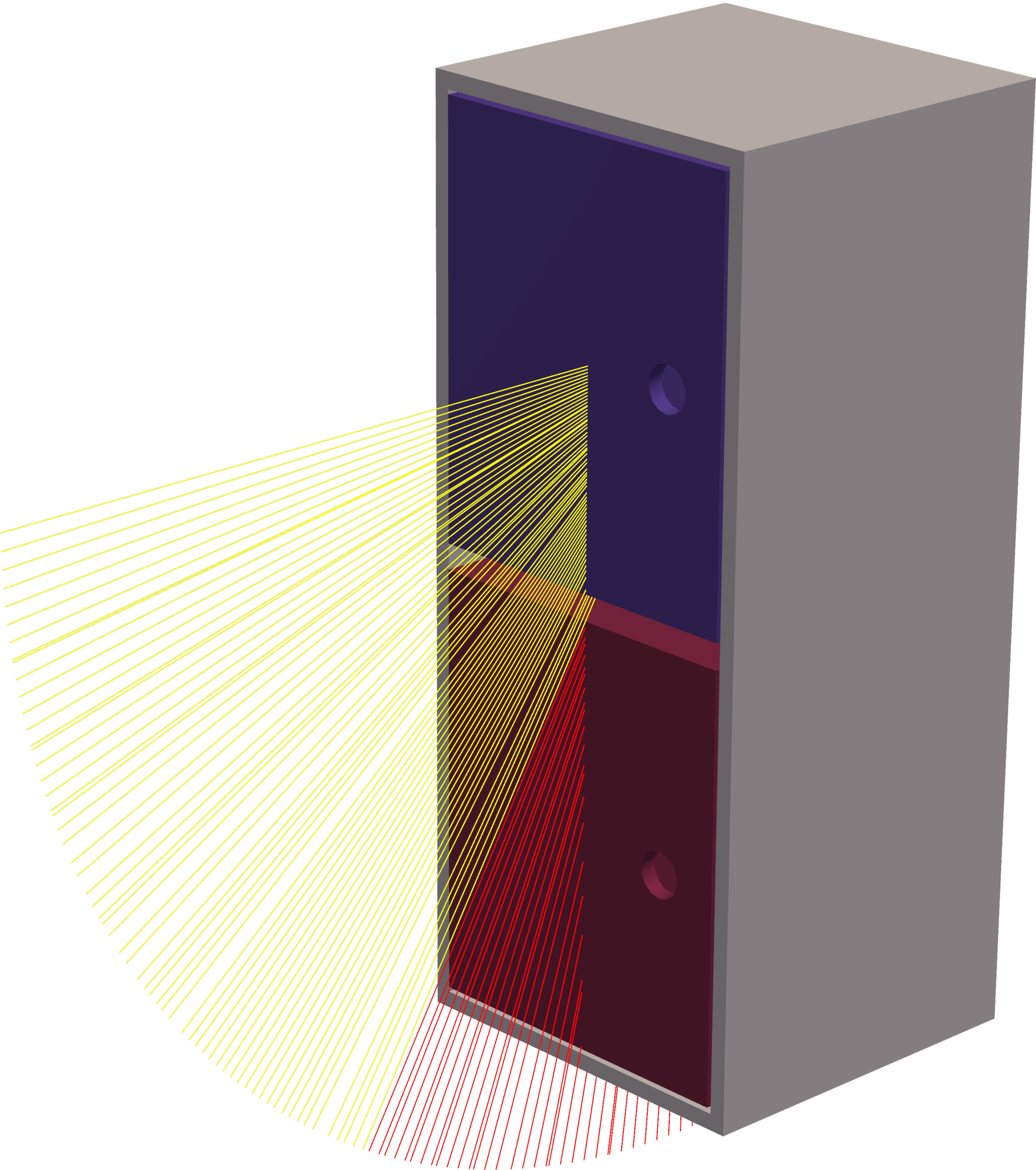}%
    \label{fig:figure_9a_vis_limit_raycast}}
    \hfil
    \subfloat[]{\includegraphics[width=.24\textwidth]{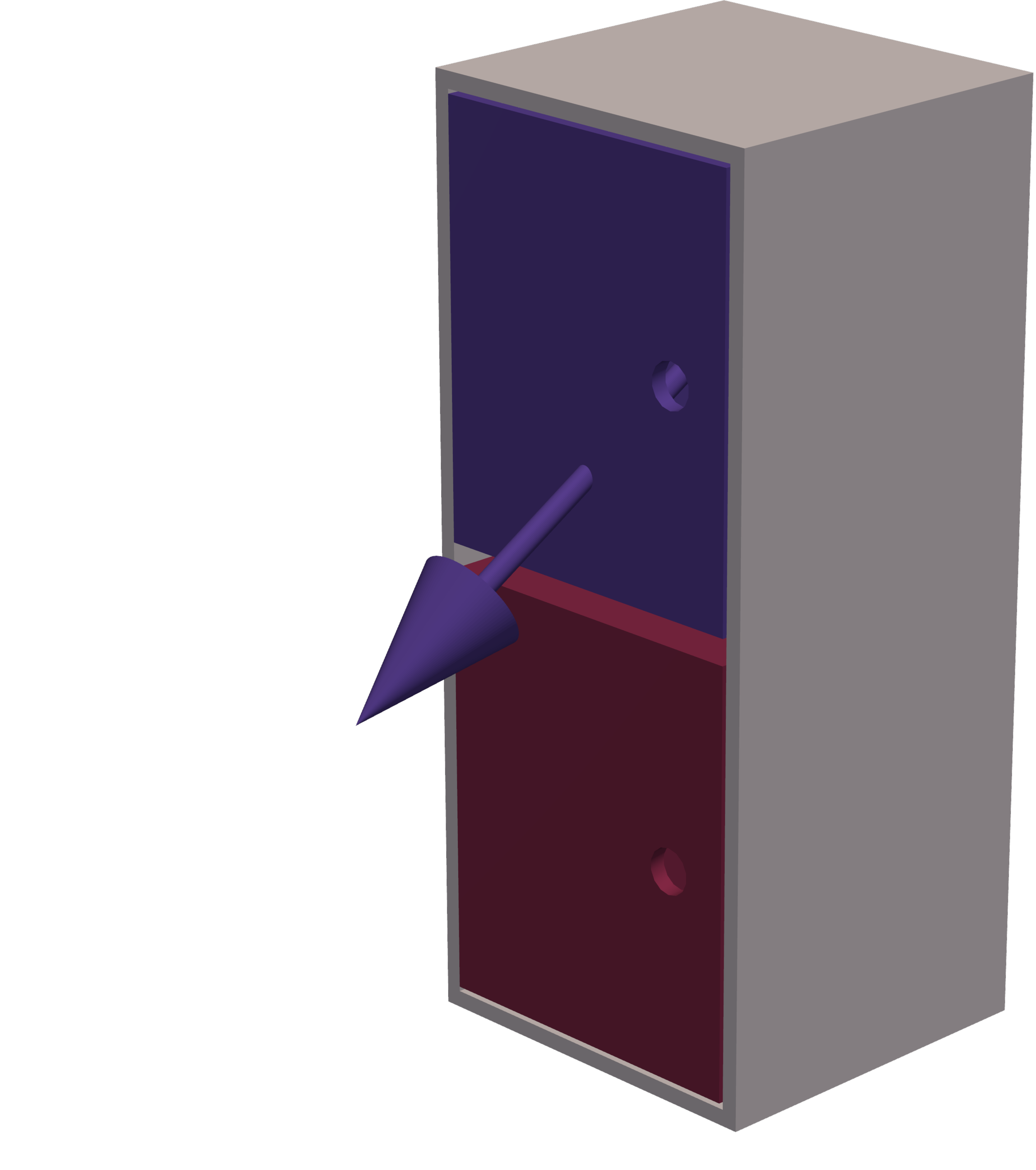}%
    \label{fig:figure_9b_vis_limit_dis_dir}}
    \hfil
    \subfloat[]{\includegraphics[width=.24\textwidth]{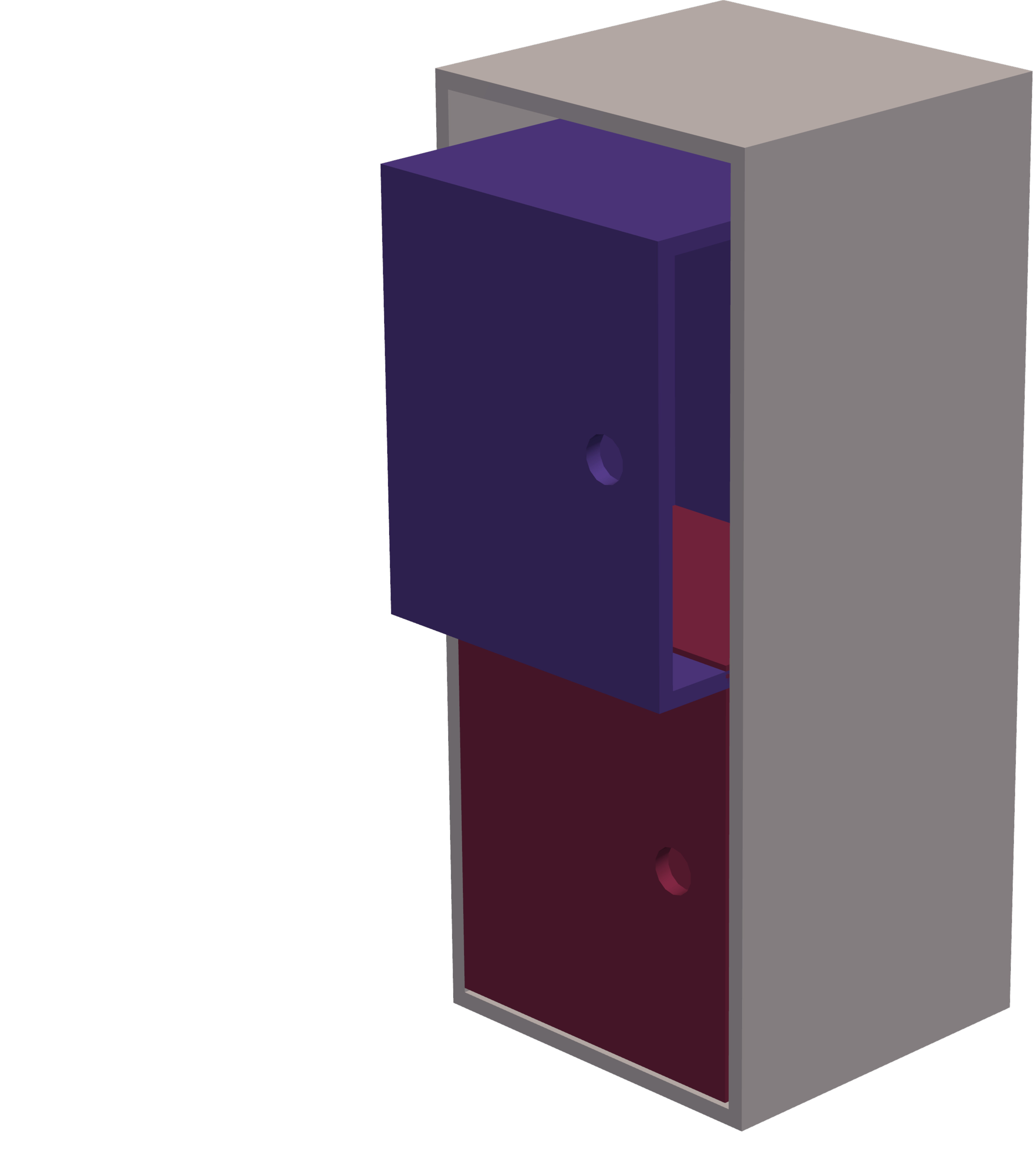}%
    \label{fig:figure_9c_vis_limit_coll}}
    \caption{Limitations of the mapping $ f_{ra}: d_{r}^{(i)} \mapsto d_{a}^{(i)} $ through ray-casting and accumulation on a simplified example of the ASAP test set. (a) shows the ray-cast over the relative disassembly space of the upper drawer (purple) with the hitting rays shown in yellow and the occluded rays shown in red. Since the two drawers are too far apart, no $congruent$ relation between them is induced. The resulting disassembly direction (b) results in a collision between the drawers (c).} 
    \label{fig:figure_9_vis_limit}
\end{figure*}

\textbf{\textit{Baseline comparison.}} To assess the practical applicability of the proposed planning framework, the planner is finally evaluated on the ASAP test dataset without the human-in-the-loop reannotation. As a baseline, we compare our results with the best-performing configuration reported by ASAP\cite{Tian2024}. As shown in \autoref{tab:ASAP_comparison}, out of 240 planning problems, our planner successfully solves 206, corresponding to a success rate of 85.83\%, which is comparable to the 82.08\% achieved by ASAP. \autoref{fig:figure_8_assembly_sequences} illustrates assembly sequences generated with our proposed framework. Some assemblies contain explicitly modeled threads or slight geometric intersections between neighboring components. For a successful disassembly, we allow for a small interference for these components. The remaining failures are primarily caused by the current limitations of our framework, namely the lack of support for rotational motions and subassembly planning, as well as assemblies that cannot be disassembled because of interlocking components. Additional failures originate from the proposed accumulation over component hitting rays used to determine possible disassembly directions. Depending on the component geometry, the shape of its relative disassembly space, and neighboring components, invalid disassembly directions may be considered. \autoref{fig:figure_9_vis_limit} shows this on a simplified example of the ASAP test set. Furthermore, the ASAP planner additionally considers assembly stability during planning, which is not yet incorporated into our framework. Despite this limitation and the additional capabilities of the ASAP baseline, the proposed planner achieves substantially lower planning times. The median runtime is reduced by more than one order of magnitude across all evaluated assembly sizes and by more than a factor of 50 for assemblies containing more than 30 components. 

Overall, the evaluation demonstrates that the proposed geometric-symbolic planning framework combines practical planning performance with high computational efficiency. Across both, the proposed dataset and the ASAP test dataset, the planner consistently scales to large industrial assemblies while maintaining planning times. Although the current implementation is limited to translational disassembly and does not yet consider stability constraints, the results indicate that the proposed geometric-symbolic formulation provides a robust and scalable foundation for autonomous assembly and disassembly planning for robotic systems.

\section{Conclusion And Outlook}

\noindent This paper presents a novel hybrid framework for ASP that combines learning-based relation extraction, human-in-the-loop annotation, and geometric-symbolic reasoning to generate executable robotic assembly and disassembly plans. PointNet++ is employed to automatically infer semantic assembly relations from point clouds, while an integrated human-in-the-loop workflow enables efficient correction of erroneous predictions and provides a mechanism to recover from planning failures. Based on the resulting symbolic assembly graph, the proposed planner derives a set of feasible manipulation primitives through explicit geometric-symbolic reasoning for the current assembly state. To improve computational efficiency, a visibility-guided search based on ray-casting is introduced to identify promising disassembly directions before collision validation. Finally, the generated set of manipulation primitives allows efficient optimization in a local search space to create disassembly sequences, enabling efficient planning for assemblies of varying size and complexity.

Although the proposed framework demonstrates promising performance, several challenges remain for future work. First, the relation extraction could benefit from larger and more balanced training dataset, particularly for underrepresented relation types. Furthermore, incorporating additional semantic or topological information may improve the discrimination between $concentric$ and $screwed$ relations, which exhibit similar geometric characteristics. Second, the current planner is limited to translational disassembly motions. Extending the proposed visibility-space formulation to full six-degree-of-freedom planning would enable rotational extraction motions and significantly broaden the range of solvable assemblies. Third, the framework currently assumes the sequential removal of individual components. Incorporating subassembly reasoning would allow groups of components to be identified and manipulated jointly. Finally, the planner focuses primarily on geometric feasibility and does not yet consider physical constraints such as gravity, stability, friction, or manipulation forces. Integrating these aspects into the symbolic planning framework would further improve the applicability of the proposed approach to autonomous manufacturing scenarios.

\section*{Funding sources}
\noindent This work was financed by the Carl-Zeiss-Stiftung under grant number P2022-01-010 and by the Federal Ministry of Research, Space and Technology (BMFTR) within the funding measure "Digital GreenTech - environmental technology meets robotics", which contributes to the BMFTR Strategy “Research for Sustainability (FONA), in the scope of the project "NewEra".

\section*{Author contributions: CRediT}
\noindent \textbf{Fabian Harlacher:} Conceptualization, Methodology, Software, Validation, Investigation, Data Curation, Writing - Original Draft, Visualization. 
\noindent \textbf{Christian Friedrich:} Conceptualization, Methodology, Writing - Original Draft, Review \& Editing, Supervision, Project administration, Funding acquisition.

\section*{Declaration of competing interests}
\noindent The authors declare that they have no known competing financial interests or personal relationships that could have appeared to influence the work reported in this paper.

\section*{Data availability}
\noindent Data will be made available on request.

\onecolumn
\appendix
\section{Dataset Overview}

\begin{table}[!hb]
\centering
\caption{Overview of the used dataset}
\begin{tabular}{@{}llll@{}}
\toprule
\multicolumn{1}{c}{\textbf{Id}} & \multicolumn{1}{c}{\textbf{Assembly Name}}                                             & \multicolumn{1}{c}{\textbf{source}}                   \\ \midrule
id\_1                            & Plate                                                                                 & own model                                             \\
id\_2                            & Joint                                                                                 & own model                                             \\
id\_3                            & Valve                                                                                 & own model                                             \\
id\_4                            & Wall-Block                                                                            & own model                                             \\
id\_5                            & Test\_bench                                                                           & own model                                             \\
id\_6                            & Block\_in\_Corner                                                                     & own model                                             \\
id\_7                            & Block\_with\_bevel                                                                    & own model                                             \\
id\_8                            & Puzzle\_with\_base\_plate\_1                                                          & own model                                             \\
id\_9                            & Puzzle\_with\_base\_plate\_2                                                          & own model                                             \\
id\_10                           & Cranfield                                                                             & \cite{CranfieldBenchmark} modified                                             \\
id\_11                           & Board 2                                                                               & \cite{luoFMBFunctionalManipulation2024}                 \\
id\_12                           & 0026a5f6c553115bfbb433eb\_19c529f969c49cfcbbfd2de7\_d473d2d63b3319b98436ab37\_default & \cite{AutoMate}                                         \\
id\_13                           & 05dfa562b007aa90fb03a03d\_764f1136b9c8b09457c02a97\_46925533868e505c79a8b722\_default & \cite{AutoMate}                                         \\
id\_14                           & 05f183912d0e14a6b76046b1\_bcaba1c3a9bb29d44f0b2216\_54209e7f2b29e4e9e9f315e7\_default & \cite{AutoMate}                                         \\
id\_15                           & 065f5b3e7a5c3419c22bb397\_080e4688978da5adf7f74ee7\_1b6a10b1552441b007a7ca27\_default & \cite{AutoMate}                                         \\
id\_16                           & 067f6b9942a1e5fd04062b43\_603f5d9d55453fcbaceb48c4\_768854bb2ac4fbdb58d13d84\_default & \cite{AutoMate}                                         \\
id\_17                           & 0611452506f549af59103205\_e97d4ad9fcd19b5f2707b0d3\_9449fe8db7fe494ed0d95540\_default & \cite{AutoMate}                                         \\
id\_18                           & 06ab4493532f949234910d00\_c19f072c571fc1df481ffd35\_3e266c5853fd4a3fee790c04\_default & \cite{AutoMate}                                         \\
id\_19                           & flange-coupling-11                                                                    & \cite{lupinettiContentbasedMulticriteriaSimilarity2019} \\
id\_20                           & flange-coupling-2                                                                     & \cite{lupinettiContentbasedMulticriteriaSimilarity2019} \\
id\_21                           & flange-coupling-21                                                                    & \cite{lupinettiContentbasedMulticriteriaSimilarity2019} \\
id\_22                           & Multiplate-F1-Clutch\_Assemble                                                        & \cite{GrabCADMakingAdditive}                            \\
id\_23                           & Minimoto\_engine                                                                      & \cite{GrabCADMakingAdditive}                            \\
id\_24                           & centrifugal-pump-41                                                                   & \cite{GrabCADMakingAdditive}                            \\
id\_25                           & 24830\_8328e407                                                                       & \cite{willis2021joinable}                               \\ \bottomrule
\end{tabular}
\label{tab:dataset_sources}
\end{table}

\clearpage
\twocolumn
\begingroup
\bibliographystyle{elsarticle-num}
\bibliography{bibliography.bib}
\endgroup
\end{document}